\documentclass[conference]{IEEEtran}
\IEEEoverridecommandlockouts
\usepackage{cite}
\usepackage{amsmath,amssymb,amsfonts}
\usepackage[noend]{algpseudocode}
\usepackage{algorithmicx,algorithm}
\usepackage{enumitem}
\usepackage{graphicx}
\usepackage{textcomp}
\usepackage{subfigure}
\usepackage{xcolor}
\usepackage{multirow}
\usepackage{booktabs} 
\usepackage{microtype}
\usepackage{color}
\usepackage{pifont}
\usepackage{tikz}

\newcommand{\circledletter}[1]{%
    \tikz[baseline=(char.base)]{%
        \node[circle, fill=black, text=white, inner sep=0.4pt] (char) {#1};%
    }%
}
\def\BibTeX{{\rm B\kern-.05em{\sc i\kern-.025em b}\kern-.08em
    T\kern-.1667em\lower.7ex\hbox{E}\kern-.125emX}}

\begin{document}

\title{SynerDiff: Synergetic Continuous Batching for Fast and Parallel Diffusion Model Inference}

\IEEEoverridecommandlockouts
\author{
	\IEEEauthorblockN{
		\textit{Ziqi Zhou}\IEEEauthorrefmark{2}, 
		\textit{Peng Yang}\IEEEauthorrefmark{2}\IEEEauthorrefmark{1},  
		\textit{Yuxin Liang}\IEEEauthorrefmark{2}, 
		\textit{Mingliu Liu}\IEEEauthorrefmark{3}, 
		and \textit{Jia Lu}\IEEEauthorrefmark{4}
    } 
	\IEEEauthorblockA{\IEEEauthorrefmark{2} School of Electronic Information and Communications, Huazhong University of Science and Technology, Wuhan, China}
	\IEEEauthorblockA{\IEEEauthorrefmark{3} The State Grid Hubei Electric Power Research Institute, Wuhan, China}
	\IEEEauthorblockA{\IEEEauthorrefmark{4} Jincheng Yunxiang Big Data Technology Operation Co., Ltd., Shanxi, China}
	\IEEEauthorblockA{Email: \{ziqi\_zhou, yangpeng, yuxinliang\}@hust.edu.cn, \{liumingliu\}@whu.edu.cn, \{jcsyxdsj\}@163.com}
    \thanks{\IEEEauthorrefmark{1}Corresponding author.}
}
\maketitle

\begin{abstract}
The expansion of Artificial Intelligence-generated content service requires diffusion model serving to simultaneously achieve high throughput and low task end-to-end (E2E) latency. However, existing continuous batching methods suffer from severe resource contention during UNet-VAE concurrency, leading to latency spikes. Furthermore, concurrent multi-task scheduling entails a trade-off between UNet throughput and VAE latency across varying scheduling strategies. To address these, we propose SynerDiff, an efficient continuous batching system built on intra-inter level synergy. At the intra-concurrency level, SynerDiff alleviates resource contention by pruning component-specific resource bottlenecks via VAE Chunking and Adaptive Skip-CFG. At the inter-concurrency level, leveraging components' differential sensitivity to scheduling granularities, a threshold-aware scheduler plans concurrent sequences and tunes intra-concurrency decisions to minimize VAE latency while maintaining UNet within high-throughput threshold. Additionally, a feedback controller dynamically adjusts this threshold based on queue loads to boost system capacity ceiling. Experimental results show that, SynerDiff improves throughput by 1.6$\times$ and decreases both average E2E and P99 tail latencies by up to 78.7\%, compared to benchmarks while guaranteeing high image fidelity.
\end{abstract}


\section{Introduction}
\label{sec:intro}

The rapid proliferation of Artificial Intelligence-Generated Content (AIGC) has revolutionized digital content creation across various domains. Central to this shift are Diffusion Models (DMs)\cite{zheng2024non}, which typically execute a sequential pipeline comprising text encoding, iterative UNet denoising\cite{rombach2022high}, and VAE decoding. While these models deliver remarkable visual fidelity\cite{kong2025distributed, li2024q}, the UNet phase consumes over 90\% of the total latency, making it the primary bottleneck. As AIGC services scale globally, a critical challenge emerges: simultaneously achieving high system throughput and low end-to-end (E2E) task latency. High concurrency often dilutes per-task resource allocation, leading to a fundamental trade-off where throughput gains come at the expense of individual response times.

Existing solutions struggle to resolve this tension effectively. Single-task optimizations\cite{ma2024deepcache, agarwal2024approximate} reduce individual latency but often rely on a single-batch paradigm that fails to exploit GPU parallel computing capabilities, resulting in insufficient system throughput\cite{gao2024Characterizing}. Conversely, although standard dynamic batching improves throughput through weight reuse, its \textit{all-in-all-out} synchronicity is ill-suited for the heterogeneous denoising steps inherent in DM requests. This mismatch triggers a severe straggler effect, forcing completed UNet tasks to wait for slower ones before proceeding to VAE decoding, which wastes GPU compute resources and unnecessarily inflates E2E latency.

\begin{figure}[t]
   \centering
           \setlength{\abovecaptionskip}{-5pt}
   \begin{minipage}[t]{0.241\textwidth}
     \centering
     \includegraphics[width=\textwidth]{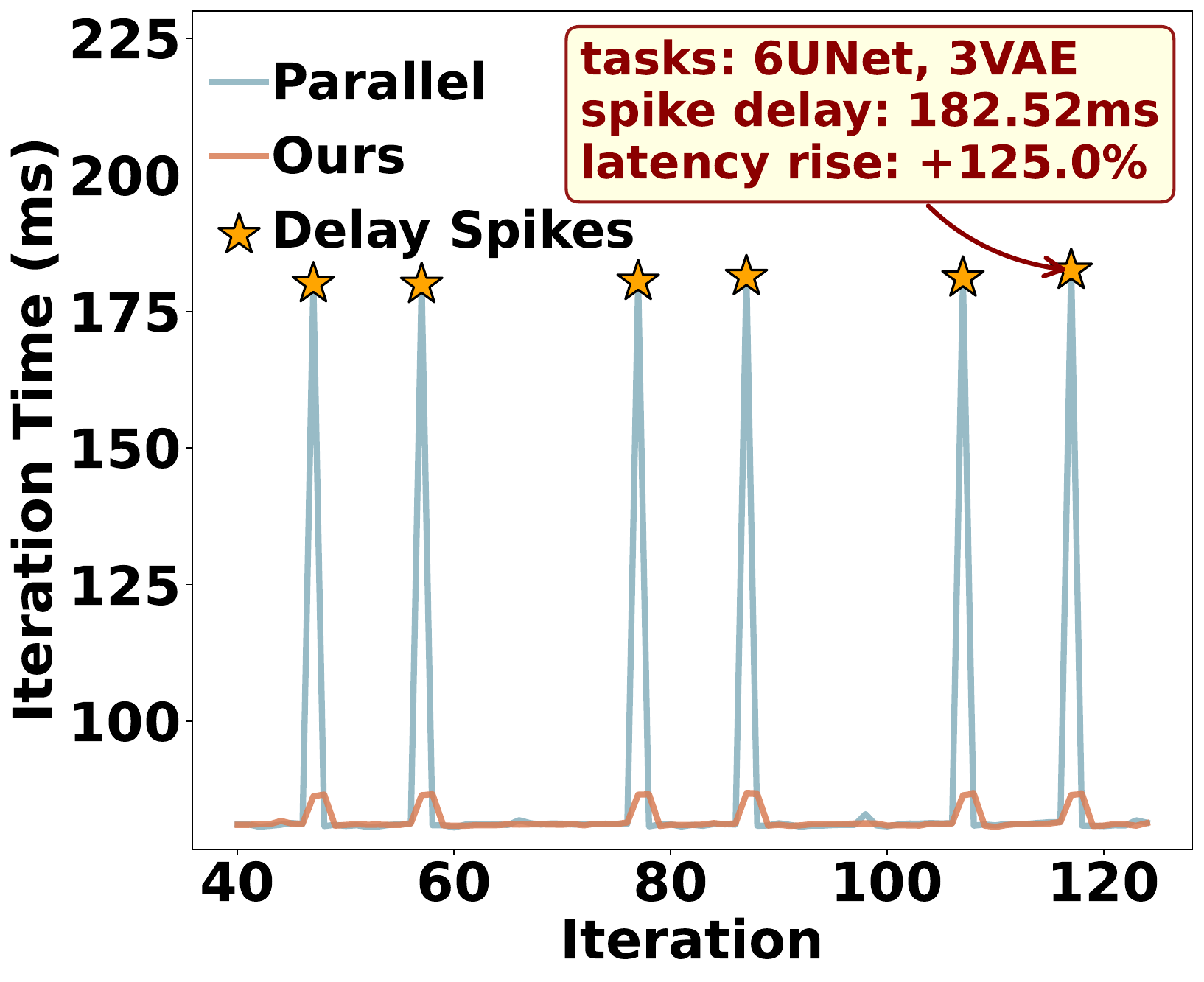}
     \caption{Latency spikes in concurrent execution of UNet and VAE.}
     \label{fig:intro1}
   \end{minipage}
   \hfill
   \begin{minipage}[t]{0.24\textwidth}
     \centering
     \includegraphics[width=\textwidth]{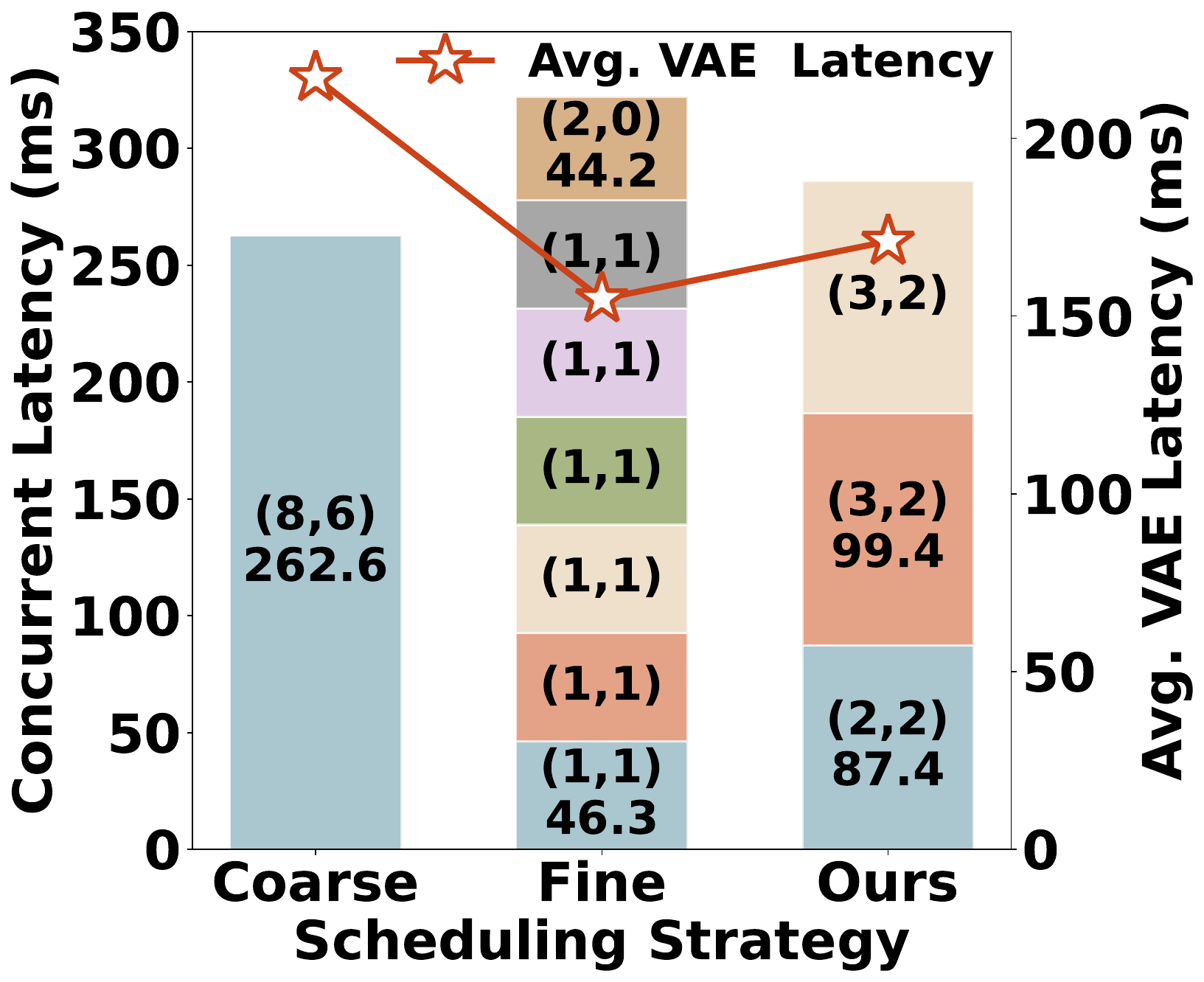}
     \caption{Trade-offs between concurrent latency and VAE latency. $(m,n)$ means $m$ and $n$ concurrent tasks}
 \label{fig:intro2}
   \end{minipage}
   	\vspace{-0.3cm}
 \end{figure}

Continuous batching \cite{agrawal2024taming} mitigates the straggler effect by decoupling diffusion model (DM) components into independent processes and adopting step-level scheduling, which enables completed UNet tasks to exit early and facilitates batch refilling. However, such fine-grained pipelining introduces a critical challenge: tasks switching from UNet denoising to VAE decoding inevitably overlap with ongoing UNet batch iterations. As depicted in Fig. \ref{fig:intro1}, such parallel execution triggers severe latency spikes, which originate from fierce resource contention between the compute-bound UNet and memory-bound VAE for limited GPU computing resources and memory bandwidth\cite{agrawal2024taming}; this issue is further analyzed in Section II. Such concurrent execution significantly inflates the latency of both tasks. Critically, these conflicts repeat at every task transition, resulting in cumulative latency spikes that reduce overall system throughput and severely degrade P99 tail latency.

Beyond the intra-concurrency resource contention, inter-concurrent multi-task scheduling granularity profoundly impacts performance. In Fig. \ref{fig:intro2}, coarse-grained strategies parallelize all pending VAE tasks with the entire UNet batch, maximizing UNet throughput via large batching but severely inflating VAE latency due to extreme resource contention. Conversely, fine-grained strategies decompose concurrency into multiple rounds of smaller task pairs; while this reduces instantaneous load and minimizes VAE latency, it diminishes system throughput by underutilizing batching benefits. This conflict roots in the asymmetric requirements of DM components: one-shot VAE decoding is highly latency-sensitive for E2E speed, while iterative UNet denoising demands high throughput to accelerate progress, leading to a cumulative global trade-off between overrall system throughput and task E2E latency.

\begin{figure}[t]
   \centering
           \setlength{\abovecaptionskip}{-5pt}
   \begin{minipage}[t]{0.243\textwidth}
     \centering
     \includegraphics[width=\textwidth]{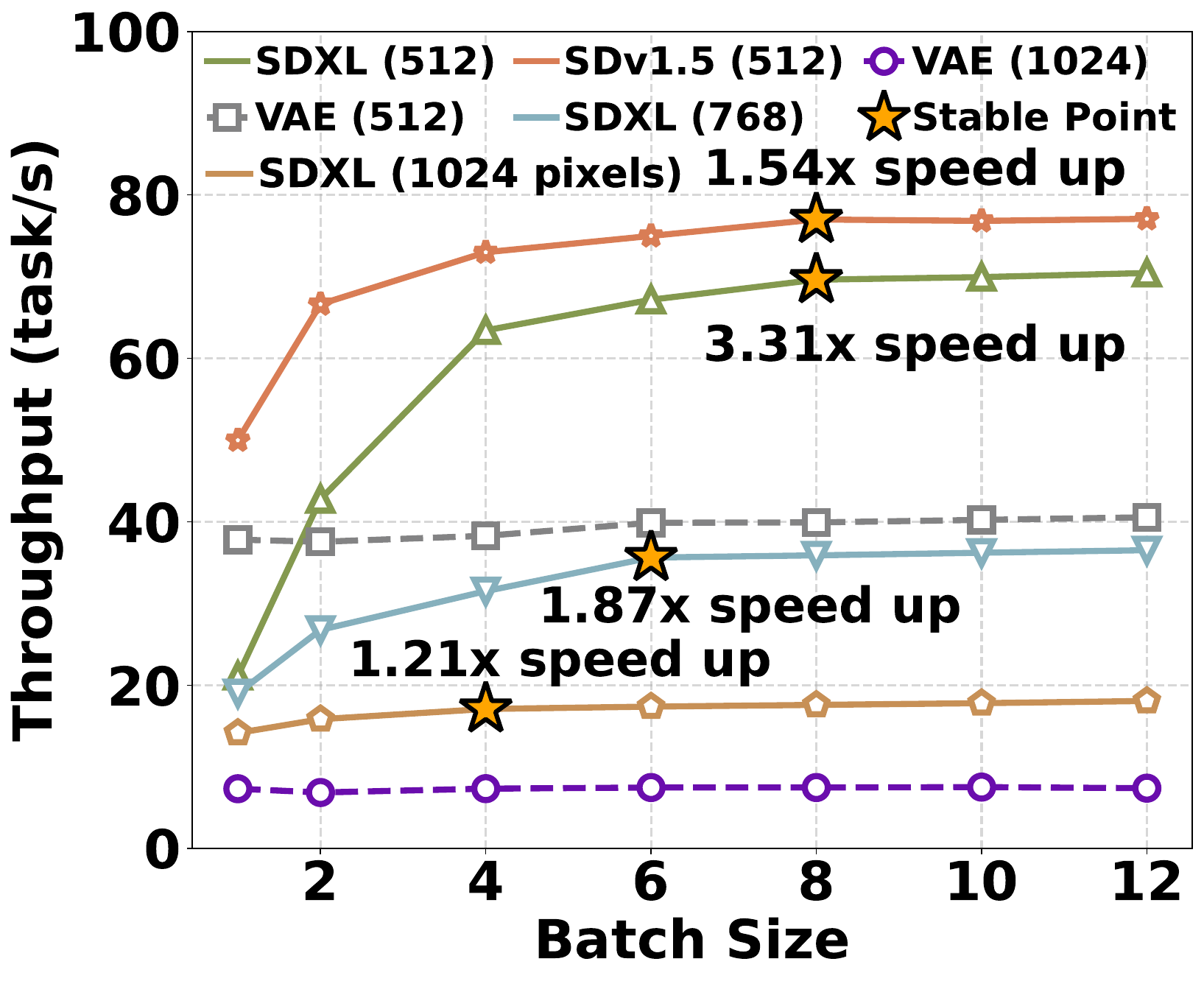}
     \caption{Throughput vs. Batch Size for UNet and VAE components.}
     \label{fig:mot1}
   \end{minipage}
   \begin{minipage}[t]{0.238\textwidth}
     \centering
     \includegraphics[width=\textwidth]{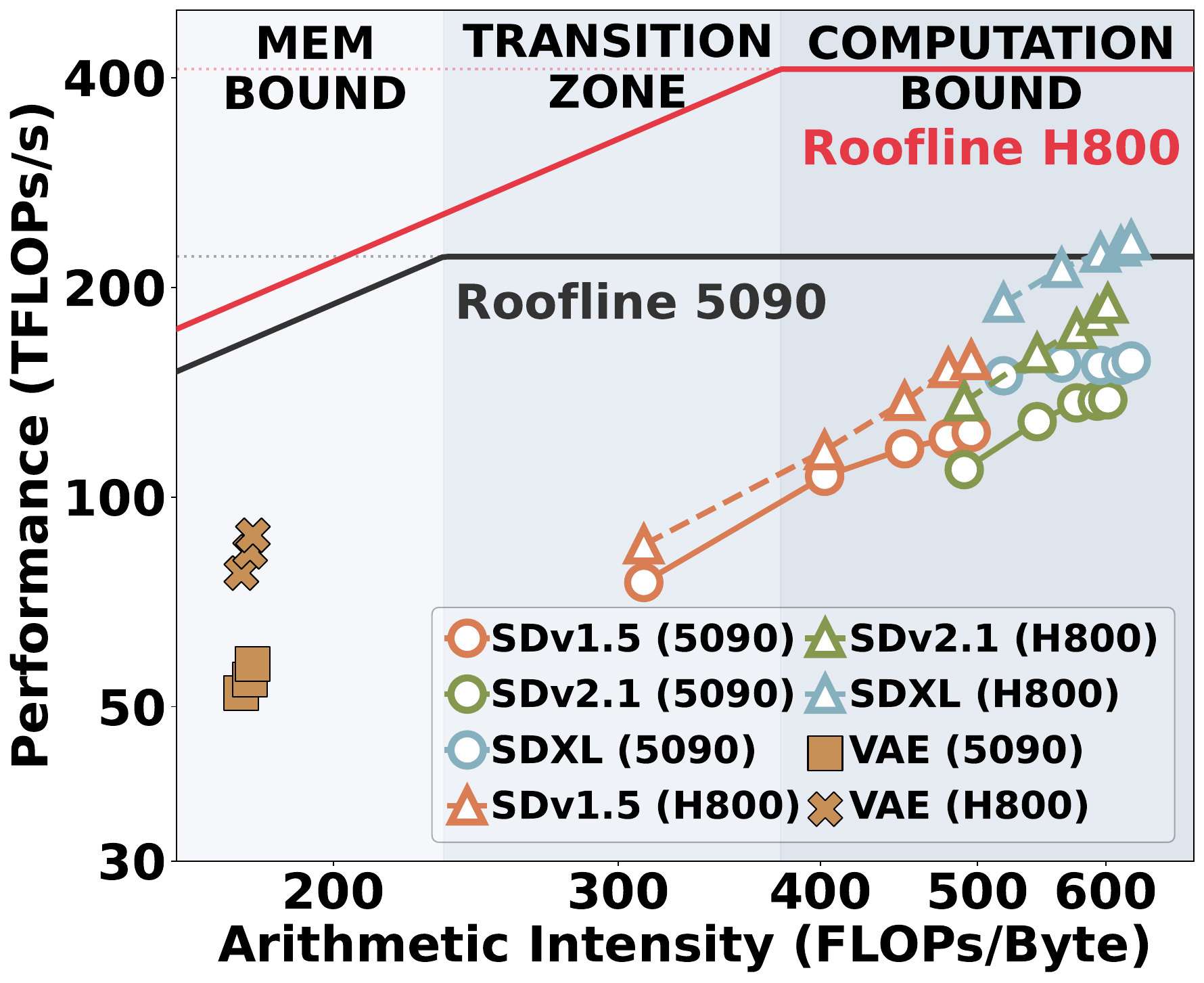}
     \caption{Roofline analysis of the UNet and VAE.}
     \label{fig:mot2}
   \end{minipage}
   \vspace{-0.3cm}
   \end{figure}

To address these challenges, we propose SynerDiff, a continuous batching system designed for high-throughput and low-latency DM serving. SynerDiff operates through intra-inter level synergy: At the intra-concurrency level, by exploiting the resource bottlenecks between compute-bound UNet and memory-bound VAE, SynerDiff suppresses latency spikes through targeted coordination. Specifically, VAE Chunking decomposes serial decoding into sub-blocks to weaken bandwidth contention within each concurrent round, while Adaptive Skip Classifier-Free Guidance (CFG) prunes UNet compute loads by skipping late-stage guidance steps with low marginal utility. These strategies enable near-lossless parallelism while maintaining high image fidelity. At the inter-concurrency level, a threshold-aware scheduler leverages the sensitivity differential between components to optimize concurrent multi-task execution sequences and tune intra-level configurations for each round. By enforcing a throughput threshold, it minimizes VAE latency while sustaining UNet performance within its high-throughput plateau. Furthermore, a feedback controller adaptively scales this threshold based on real-time queue loads to elevate system capacity under heavy traffic\cite{kong2024adaptive}. By integrating task sequences planning with fine-grained strategy tuning, SynerDiff achieves globally optimized throughput and E2E latency. Our main contributions are summarized as follows:

\begin{itemize}
    \item We characterize the resource contention between compute-bound UNet and memory-bound VAE, identifying this contention as the cause of concurrent latency spikes. 
    \item We propose intra-concurrency strategies, including VAE Chunking and adaptive Skip-CFG to suppress resource interference by decomposing serial VAE decoding and pruning redundant compute loads of UNet denoising.
    \item We design a inter-concurrency threshold-aware scheduler that plans concurrent sequences and tunes intra-level decisions to minimize E2E latency under high-throughput threshold, further enhanced by a feedback controller.
    \item Extensive evaluations show that SynerDiff improves throughput by 1.6$\times$ and reduces average E2E and P99 latencies by up to 78.7\% with high image fidelity.
\end{itemize}

\section{Motivation}

\subsection{Profiling Concurrency Bottlenecks of UNet and VAE}


Inference latency is determined by compute capability and memory bandwidth\cite{agrawal2024taming}. To trace the root of latency spikes, we profile the throughput of UNet and VAE under varying batch sizes and resolutions in Fig. \ref{fig:mot1}. UNet shows sub-linear throughput scaling, where batching improves compute utilization. However, this scalability is resolution-dependent; for SDXL, the speedup drops from $3.31\times$ (512px) to $1.21\times$ (1024px) as larger workloads saturate hardware, indicating that UNet denoising is bounded by peak compute capacity. Conversely, VAE decoding is batch-insensitive, with throughput saturating at a single task across resolutions, revealing a memory bandwidth bottleneck that negates batching benefits.

Roofline analysis on RTX 5090 and H800 further validates these heterogeneous bottlenecks. As shown in Fig. \ref{fig:mot2}, VAE consistently lies in the memory-bound region across resolutions and platforms, driven by massive high-resolution activations in convolutional layers that saturate memory bandwidth\cite{liang2025networked}. Conversely, UNet variants are compute-bound, where batching increases Arithmetic Intensity (AI) via weight amortization until architectural saturation, yet its performance still falls below the theoretical peak, leaving residual compute headroom. Given their complementary resource demands, co-execution of UNet and VAE is promising to fully exploit both compute and bandwidth efficiency. However, naive parallelism triggers severe latency spikes, since the compute overhead of VAE’s high-resolution operations exceeds spare GPU capacity, and its intensive bandwidth usage conflicts with UNet memory access, causing fierce cross-resource contention. Such coarse-grained contention highlights the demand for fine-grained synergetic optimizations to address these heterogeneous bottlenecks.

\subsection{Exploring Fine-grained Concurrency Coordination}

\begin{figure}[t]
	\centering
            \setlength{\abovecaptionskip}{-6pt}
	\includegraphics[width=1\linewidth]{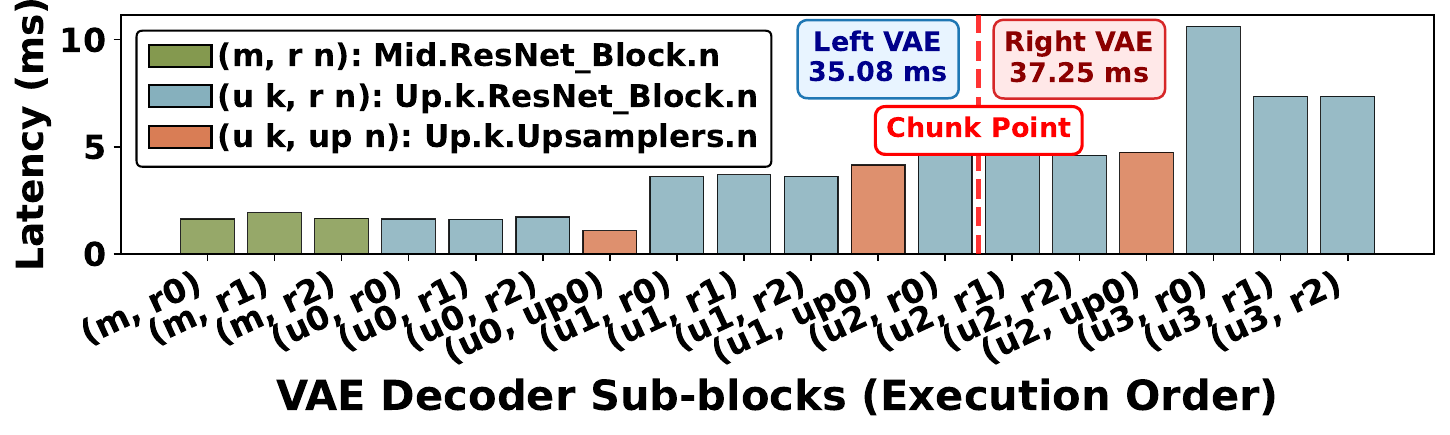}
	\caption{Structural decomposition of the VAE decoder.}
	 \label{fig:mot3}
   	\vspace{-0.3cm}
\end{figure}

The persistent bandwidth occupation during VAE decoding stems from the generation of high-dimensional intermediate activations. Executing a full VAE task concurrently with UNet batches inevitably starves the memory operations critical for denoising. Recognizing that the VAE decoder is structurally dominated by serial ResNet Blocks shown in Fig. \ref{fig:mot3}, we partition it into sub-blocks to enable fine-grained concurrency. This structural decomposition fragments continuous, high-intensity bandwidth demand into discrete, shorter intervals, temporally dispersing pressure to preserve memory headroom for UNet and mitigate transient contention. However, VAE Chunking inherently trades off individual decoding latency by extending execution across multiple rounds. While bandwidth coordination is essential, UNet inherently dictates the computational boundary during concurrency. The standard CFG mechanism exacerbates this compute footprint by simultaneously evaluating both conditional and unconditional branches. However, This architecture contains exploitable computational redundancy, selectively bypassing unconditional evaluations can generate substantial compute slack without disrupting denoising. This insight motivates a workload pruning approach to compress UNet's instantaneous compute demand, carving out necessary scheduling headroom for VAE decoding. Consequently, coupling bandwidth coordination with workload pruning emerges as the essential intra-concurrency paradigm for resolving the inherent resource bottlenecks of UNet-VAE concurrency.

\begin{table}[t]
  \centering
  \small 
  \caption{\textnormal{Performance under varying concurrency granularities. $(x, y)$ indicates sequential rounds of $x$ and $y$ concurrent UNet-VAE pair tasks.}} 
  \label{tab:1}
  \vspace{-0.2cm} 
  
  \resizebox{\columnwidth}{!}{
    \renewcommand{\arraystretch}{1.1} 
    \begin{tabular}{c | c c c c c c}
    \toprule
    \textbf{Scheduler} & \textbf{(6)} & \textbf{(1, 5)} & \textbf{(3, 3)} & \textbf{(1, 1, 4)} & \textbf{(1, 2, 3)} & \textbf{(1, 1, 1, 1, 1, 1)} \\ \hline
    UNet Latency (ms) & \textbf{237.9} & 250.2 & 249.0 & 252.7 & 256.2 & 277.8 \\ \hline
    VAE Latency (ms)  & 204.2 & 188.6 & 173.3 & 161.6 & 145.3 & \textbf{124.8} \\ 
    \bottomrule
    \end{tabular} 
  }
  \vspace{-0.3cm} 
\end{table}

\subsection{Non-linear Throughput Plateaus and Scheduling Insights}
Beyond latency spikes, multi-task scheduling involves inter-concurrency level latency trade-offs across concurrency granularities. The results in Table \ref{tab:1} reveal a striking performance asymmetry: while fine-grained scheduling steadily reduces average VAE latency, UNet performance follows a non-linear \textit{plateau-then-spike} pattern. This plateau arises from the sub-linear scaling of UNet throughput shown in Fig. \ref{fig:mot1}, where increasing concurrency granularity within the medium range has minimal marginal impact on throughput. This observation suggests that effective scheduling should prioritize the optimization of VAE latency within the UNet’s plateau region.

\begin{figure}[t]
	\centering
            \setlength{\abovecaptionskip}{-3pt}
	\includegraphics[width=1\linewidth]{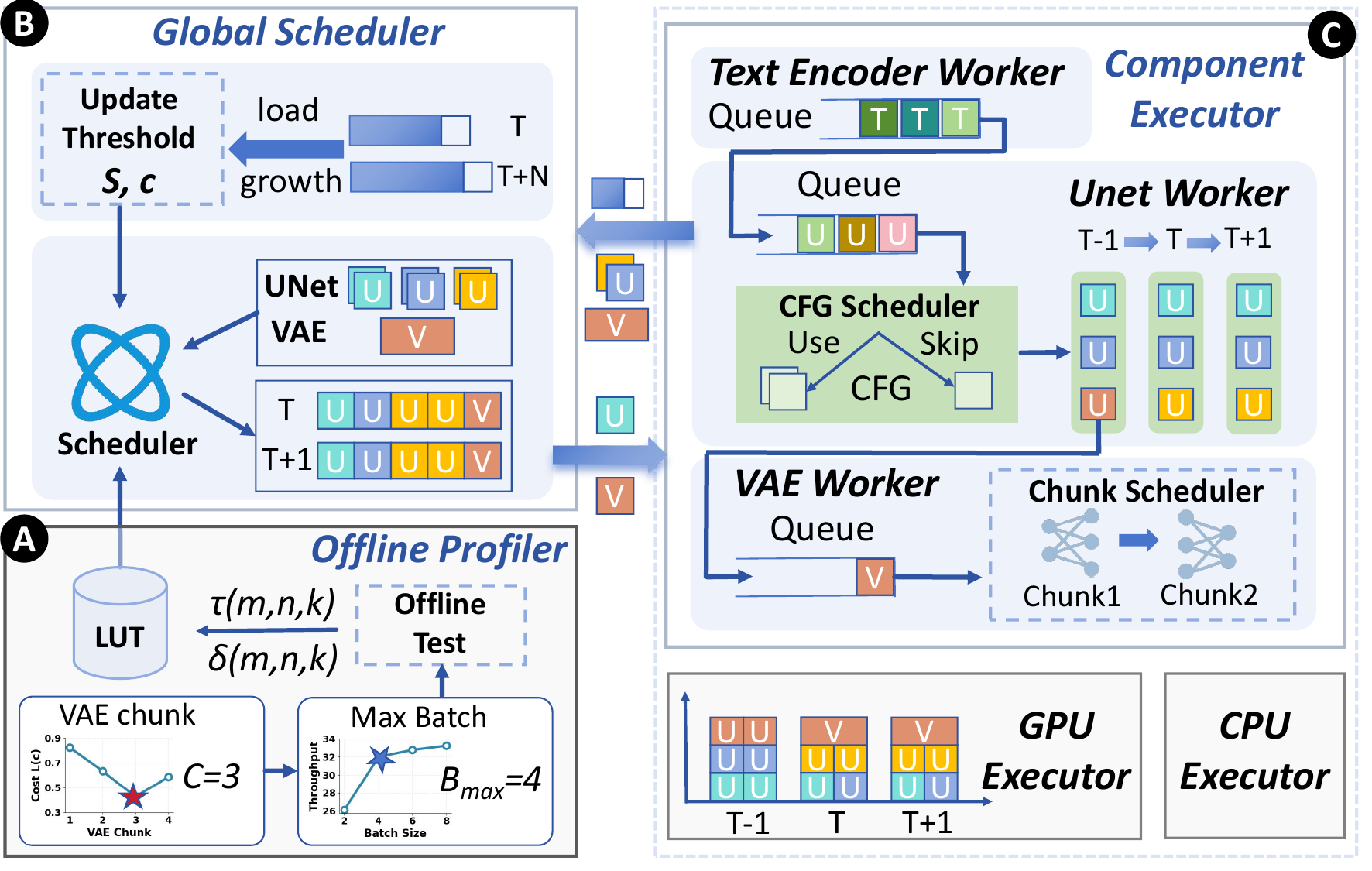}
	\caption{ The workflow of SynerDiff.}
	\label{fig:workflow}
   	\vspace{-0.3cm}
\end{figure}

\section{System Design}

\subsection{System Overview}

SynerDiff is a component-synergetic continuous batching system for fast and parallel DM serving, It optimizes both system throughput and task end-to-end latency through intra-concurrency coordination with VAE Chunking and adaptive Skip-CFG, and inter-concurrency scheduling via threshold-aware sequencing. As illustrated in Fig. \ref{fig:workflow}, SynerDiff comprises three core components. \circledletter{A} \textbf{Offline Profiler} conducts micro-benchmarks to determine static configurations, such as VAE Chunking granularity and maximum batch limits $B_{max}$, and constructs a concurrent-latency lookup table to guide runtime optimization. \circledletter{B} \textbf{Global Scheduler} is the control plane, acting as the centralized controller that monitors system states, including queue workloads and batching progress. It performs threshold-aware sequences scheduling and dynamically adjusts throughput thresholds based on real-time traffic. And \circledletter{C} \textbf{Component Executor} is the data plane. It manages fine-grained pipeline execution and inter-stage communication, processing micro-batches and executing intra-concurrency schedules, such as Skip-CFG and VAE Chunking, as directed by the control plane. Upon arrival, tasks follow a pipeline of CPU-side text encoding, GPU-based UNet denoising and VAE decoding. The global scheduler orchestrates concurrency by assigning execution sequences, Skip-CFG statuses, and VAE Chunking granularities to be implemented for precise local execution. Finally, the proposed SynerDiff architecture is model-agnostic and applicable to various UNet-based DM serving systems.


\subsection{Intra-Concurrency Coordination}
\begin{figure}[t]
	\centering
            \setlength{\abovecaptionskip}{-3pt}
	\includegraphics[width=1\linewidth]{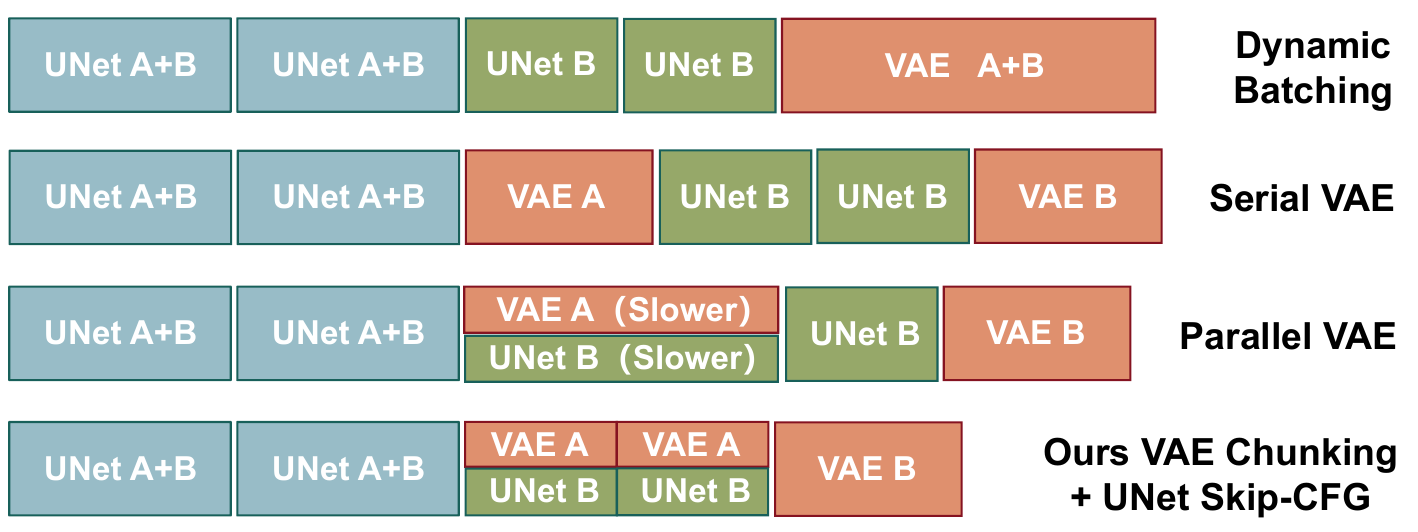}
	\caption{Comparative analysis of concurrent scheduling strategies.}
	\label{fig:system1}
   	\vspace{-0.3cm}
\end{figure}

\begin{figure}[t]
   \centering
           \setlength{\abovecaptionskip}{-3pt}
   \begin{minipage}[t]{0.243\textwidth}
     \centering
     \includegraphics[width=\textwidth]{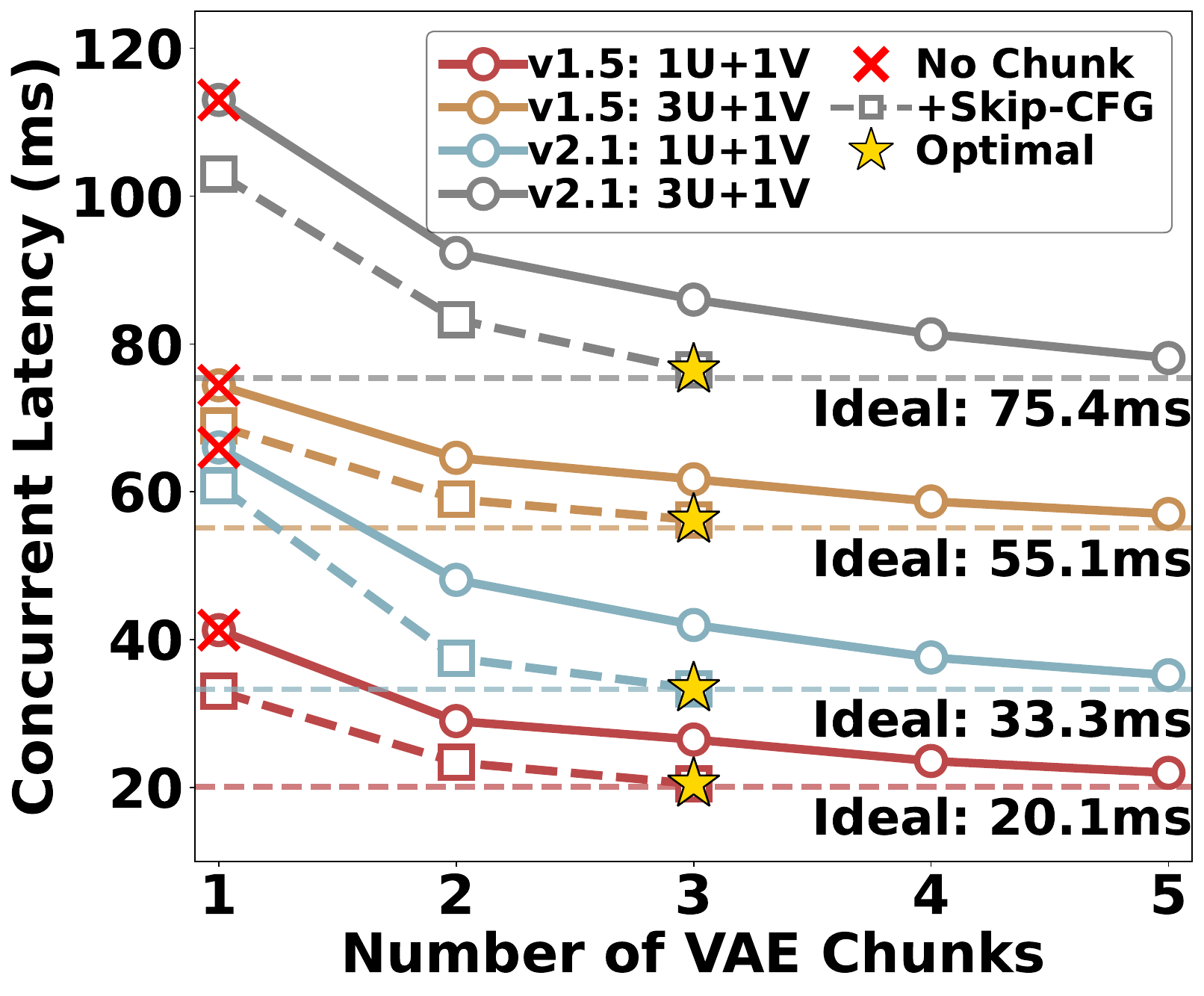}
     \caption{Impact of VAE Chunking and Skip-CFG on concurrent latency.}
     \label{fig:system2}
   \end{minipage}
   \begin{minipage}[t]{0.238\textwidth}
     \centering
     \includegraphics[width=\textwidth]{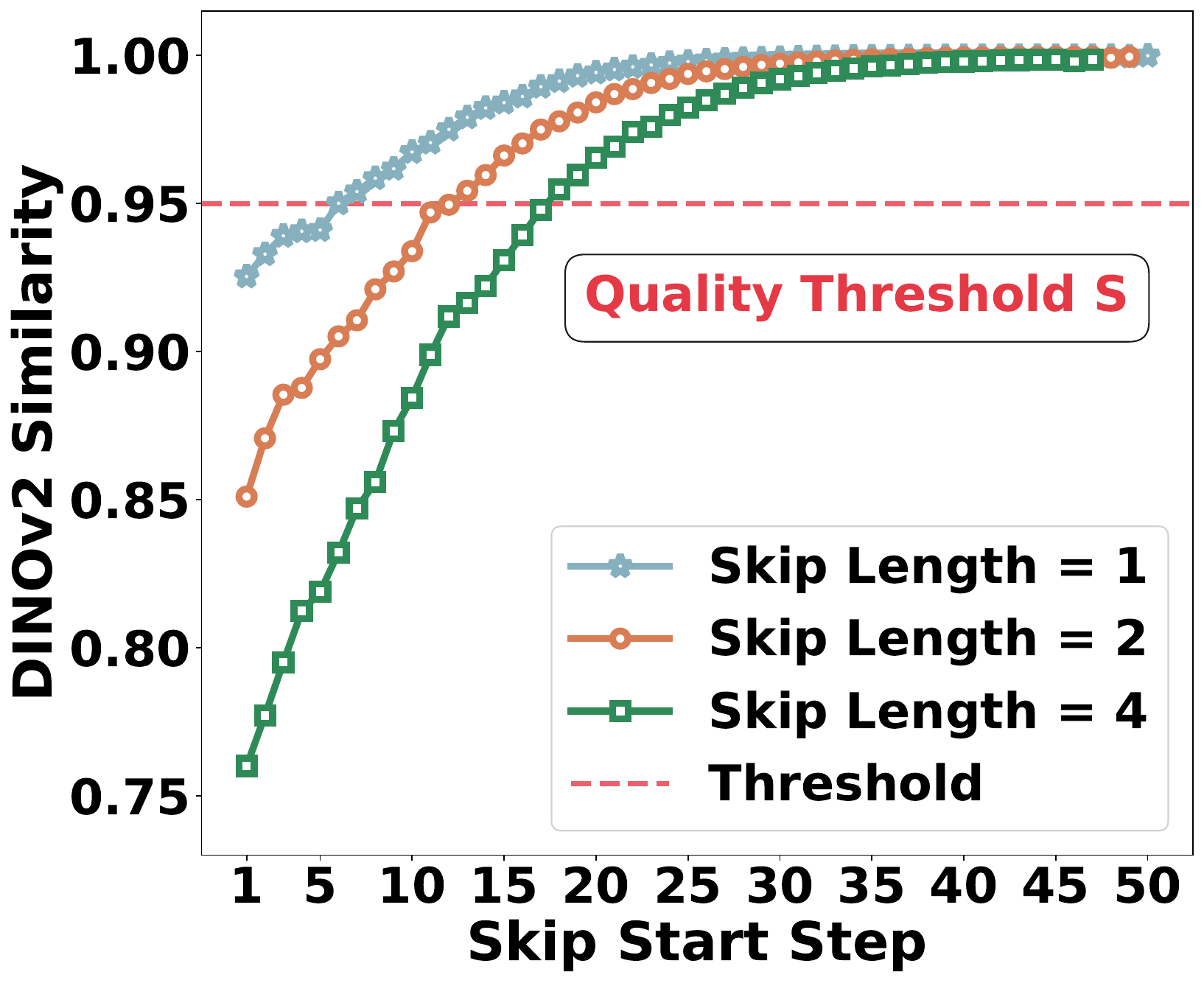}
     \caption{Impact of Skip-CFG on generated image fidelity.}
     \label{fig:system3}
   \end{minipage}
   \vspace{-0.3cm}
 \end{figure}

To mitigate latency spikes caused by severe resource contention between concurrent components, SynerDiff employs VAE Chunking with UNet workload pruning as the intra-concurrency strategy. VAE Chunking partitions the decoder into $N$ temporally equivalent sub-blocks at the ResNet block level. This boundary-aligned decomposition avoids additional kernel launch overhead and suppresses peak memory bandwidth occupancy via interleaved scheduling with UNet within each concurrent round. Concurrently, SynerDiff utilizes Adaptive Skip-CFG for UNet workload pruning, eliminating unconditional branches to reduce computational redundancy during denoising. As shown in Fig. \ref{fig:system1}, upon UNet-VAE concurrency, while serial scheduling incurs pipeline stalls and naive parallelism triggers severe resource contention, SynerDiff’s fine-grained interleaving, combined with Skip-CFG workload reduction, disrupts the VAE’s bandwidth monopoly and alleviates compute pressure without interrupting UNet denoising progress. As shown in Fig. \ref{fig:system2}, this strategy effectively suppresses latency spikes, and optimal configurations achieve near-lossless parallelism, particularly during high-resolution generation (\textit{e.g.}, SDv2.1), where intensified compute and bandwidth demands necessitate Skip-CFG and VAE Chunking. 

The impact of Skip-CFG on image fidelity is evaluated via DINO similarity analysis\cite{oquab2023dinov2}. Fig. \ref{fig:system3} reveals that early denoising stages are highly CFG-sensitive, where brief skipping significantly degrades quality. Conversely, intermittent Skip-CFG (\textit{e.g.}, 1–2 steps) during later stages leads to negligible impact. This late-stage error tolerance enables workload reduction without compromising generation quality \cite{sheng2025text}. Consequently, SynerDiff introduces a quality threshold $S$ to constrain the Skip-CFG execution window and preserve fidelity. However, optimizing these strategies involves a complex design space: excessive Chunking inflates decoding latency, while Skip-CFG entails a fidelity-throughput trade-off under heavy workloads. Furthermore, multi-task scenarios require optimizing multi-round concurrent execution sequences. To address this, SynerDiff integrates offline profiling with online multi-task scheduling to synergistically optimize both inter-concurrency sequencing and intra-concurrency decision-making.

\subsection{Offline Profiling}
The offline phase determines the optimal VAE Chunking granularity and peak UNet batch size $B_{max}$ tailored to the platform and model, then it constructs a concurrent latency lookup table with various UNet-VAE combinations. First, we define $C_{max}$ as the maximum factor limiting latency overhead to within 5\% of solo UNet execution. Selecting VAE Chunking granularity $c \in \{1, \dots, C_{max}\}$ involves a core trade-off. While finer Chunking minimizes concurrent interference, it extends VAE latency as the progression of sequential sub-blocks is bottlenecked by concurrent UNet steps. We derive the optimal baseline factor $c$ by minimizing a weighted cost function $L(c)$:
\begin{equation}
\small
\setlength\abovedisplayskip{3pt}
\setlength\belowdisplayskip{3pt}
\min L(c) = \lambda \left( \frac{T_u(c) - T_{u0}}{T_{u0}} \right) + (1 - \lambda)  \left( \frac{T_v(c) - T_{v0}}{T_{v0}} \right),
\end{equation}
where the first and second terms quantify the concurrent latency overhead for UNet and VAE, respectively. We introduce a weighting factor $\lambda \in [0, 1]$ to balance the contribution of each term. The cumulative latencies over $c$ rounds are expressed as:
\begin{equation}
\small
\setlength\abovedisplayskip{1.5pt}
\setlength\belowdisplayskip{1.5pt}
T_u(c) = \sum_{i=1}^c \max\left(t_u^i, t_v^i\right), T_v(c) = \sum_{i=1}^{c-1} \max\left(t_u^i, t_v^i\right) + t_v^c,
\end{equation}
where $t_{u}^{i}$ and $t_{v}^{i}$ are UNet and VAE latencies in round $i$. The discrepancy between $T_u(c)$ and $T_v(c)$ arises as VAE does not wait for UNet completion in the final round, being offloaded for CPU post-processing. The offline-optimized $c$ provides an initial value, which the global scheduler dynamically adjusts based on real-time queue loads. Then, we identify the saturation batch size $B_{max}$ based on the sub-linear scaling of throughput in Section II. Lastly, we build a performance lookup table $T_{cost}^{c}(m, n, k)$ within the bounds of $B_{max}$ and $C_{max}$. Here, $m, n$, and $k$ denote the total counts of UNet, VAE, and Skip-CFG-enabled UNet tasks, respectively. The table stores the profiled concurrent latency $\tau_{m,n,k}$ and VAE decoding latency $\delta_{m,n,k}$, enabling $O(1)$ runtime retrieval for online scheduling.


 \subsection{Inter-Concurrency Scheduling}In online serving, the inter-concurrency of multi-task necessitates a trade-off between UNet throughput and VAE latency. Guided by Section II-C, the global scheduler exploits the sensitivity differential between components by prioritizing VAE latency reduction within the UNet’s throughput plateau, where fine-grained execution minimizes VAE delay without sacrificing UNet performance. Simultaneously, optimizing intra-concurrency decision for each concurrent round is necessary to mitigate latency spikes. To this end, we formulate a sequence planning problem and design a 3D dynamic programming algorithm to attain a globally optimal solution that jointly determines the inter-concurrency level execution sequence and the intra-concurrency level decisions for each concurrent round.
 

For a concurrent workload of $M$ UNet and $N$ VAE tasks, let $s_{i}$ denotes the denoising progress of UNet task $u_{i}$. Given a quality threshold $S$, we derive the minimum denoising step $s_{min}$ for Skip-CFG execution based on Fig. \ref{fig:system3}; a UNet task $u_{i}$ is eligible for Skip-CFG only if $s_{i} \ge s_{min}$, thereby defining $K$ as the maximum number of such eligible tasks. For each task arriving at $A_{i}$, let $U_{i}$ and $V_{i}$ represent its completion times for UNet denoising and VAE decoding, respectively. The scheduler divides concurrent execution into $T$ discrete stages. In stage $t$, a configuration $(m_{t}, n_{t}, k_{t})$ is executed, where $m_{t}$, $n_{t}$ and $k_t$ denote the total counts of UNet, VAE, and Skip-CFG-enabled UNet tasks. We define the execution sequence as $S = [(m_{1}, n_{1}, k_{1}), \dots, (m_{T}, n_{T}, k_{T})]$ and the task mapping as $E = [E_{1}, \dots, E_{T}]$. To minimize the average E2E latency of VAE tasks under throughput and quality constraints, we formulate  the problem as a constrained sequence optimization problem to derive the optimal execution path $(S, E)$:

\begin{subequations}\label{eq:optimization}
\setlength{\jot}{-1pt} 
\begin{align}
\tiny
\setlength\abovedisplayskip{-3pt}
\setlength\belowdisplayskip{-3pt}
\mathbf{P}: &\min_{S, E} \quad  J = \frac{1}{N} \sum_{t=1}^T \sum_{i \in E_t} \left(U_i + V(t) - A_i\right) \label{eq:p} \\
\text{with} \quad & V(t) = \sum_{j=0}^{t-1} \tau(m_j, n_j, k_j) + \delta(m_t, n_t, k_t) \label{eq:def_c} \\
\text{s.t.} \quad & \sum_{t=1}^T m_t = M, \quad \sum_{t=1}^T n_t = N, \label{eq:c1} \\
& 0 \leq n_t \leq m_t, 0 \leq k_t \leq m_t, \forall t \in \{1,...,T\}, \label{eq:c2} \\
& \sum_{t=1}^T k_t \leq K, \label{eq:c3} \\
& \sum_{t=0}^T \tau(m_t, n_t, k_t) \leq (1+\alpha)\tau(M,N,K), \label{eq:c4}
\end{align}
\end{subequations}
where Eq. \eqref{eq:def_c} captures the total processing latency for $n_{t}$ concurrent VAE tasks in stage $t$, comprising both queuing and decoding delays. Eq \eqref{eq:c1}-\eqref{eq:c4} respectively ensure the completeness of task execution, adherence to per-stage task boundaries $(M,N)$, image quality preservation via Skip-CFG budgets, and throughput threshold through total latency capping. The throughput threshold is set according to the UNet throughput plateau identified in Section II-C.

The optimal substructure of problem $\mathbf{P}$ enables our 3D-DP algorithm to derive the optimal concurrent execution path $(S, E)$. As shown in Alg. \ref{alg:dp_schedule}, the state $DP[i, j, u]$ represents the minimum cumulative E2E latency for completing $i$ UNet, $j$ VAE, and $u$ Skip-CFG tasks. The algorithm explores valid actions $(m, n, k)$ to search the optimal path from $(0,0,0)$ to $(M, N, K)$, and prunes throughput-infeasible transitions to accelerate convergence. Despite its $O(M N K)$ complexity, the scheduler only introduces millisecond-level overhead in practice due to $B_{max}$ and finite queue depths, thus satisfying real-time decision requirements. For dynamic workloads, SynerDiff uses a load-adaptive feedback controller that monitors queue trends. Upon detecting queue buildup, the controller raises the quality threshold $S$ to increase Skip-CFG frequency and boost throughput, while restoring it as load subsides to prioritize image fidelity. If throughput remains insufficient even at the minimum quality threshold $S_{min}$, the controller increments VAE Chunking granularity $c$ to further mitigate hardware contention and lift the system throughput ceiling.

\section{Performance Evaluation}
\subsection{Experimental Settings}
\subsubsection{Implementation}
We evaluate SynerDiff on an NVIDIA RTX 5090 GPU featuring 32GB of memory and 231 TFLOPS of FP16 precision. While we employ SDv1.5 model for our experiments, the design of SynerDiff is inherently generalizable to other diffusion model services utilizing UNet-based denoising frameworks. To simulate realistic workloads, we sample 5,000 requests from DiffusionDB\cite{wang2023diffusiondb}, applying four kinds of Poisson arrival rates: Low, Medium, High, and Extreme. A burst traffic scenario is also included by dispatching 50\% of total tasks within a concentrated short interval. Each request generates images at discrete resolutions of 512, 768, and 1024 px, with denoising steps ranging from 20 to 50 to represent heterogeneous inference demands. For offline selection of optimal VAE Chunking granularity, we empirically set $\lambda = 0.5$. For throughput threshold of online scheduling, we set $\alpha=10\%$.
\subsubsection{Baselines}
We compare SynerDiff with following baselines:  
    \begin{itemize}
        \item \textbf{Diffusers}: A standard baseline fixed at $BS=1$ as it lacks support for batching heterogeneous denoising steps.
        \item \textbf{Dynamic Batching}: A task-level batching approach that assembles tasks within a 0.5s collection window and requires synchronous batch release.
        \item \textbf{InstGenIE}\cite{jiang2025instgenie}: The first continuous batching system for DMs, using direct UNet-VAE concurrency strategy.
    \end{itemize}

For a fair comparison, the maximum batch size is set to $BS=8$ for SynerDiff, InstGenIE, and Dynamic Batching.
\subsubsection{Metrics}System efficiency is evaluated via throughput, average E2E latency, and P99 tail latency. Image quality is quantified using DINO \cite{oquab2023dinov2} and CLIP scores \cite{hessel2021clipscore}, representing visual fidelity and semantic text-image alignment, respectively.
 
\begin{algorithm}[t]
\caption{Optimal Sequence Selection via 3D-DP}\label{alg:dp_schedule}
    \hspace*{0.02in}{\bf Input:} $M, N, K$, Latency $\tau, \delta$, Throughput limit $T_{lim}$\\
    \hspace*{0.02in}{\bf Output:} Optimal sequence $\mathcal{S}$
    \begin{algorithmic}[1]
    \State $DP[M][N][K] \leftarrow \langle \infty, 0 \rangle$ \Comment{$\langle c:\text{Cost}, t:\text{Time} \rangle$}
\State $DP[0][0][0] \leftarrow \langle 0, 0 \rangle$
\For{state $(i, j, u)$; $DP[i, j, u].c < \infty$}
\For{action $(m, n, k)$; $i+m \le M, j+n \le N, u+k \le K$}
\State $T_{new} \leftarrow DP[i, j, u].t + \tau(m, n, k)$
\If{$T_{new} \le T_{lim}$}
\State $C_{v} \leftarrow (n > 0) ? (DP[i, j, u].t + \delta(m, n, k)) : 0$ 
\State $Cost_{new} \leftarrow DP[i, j, u].c + C_{v} \cdot n$
\If{$Cost_{new} < DP[i+m, j+n, u+k].c$}
\State $DP[i+m, j+n, u+k] \leftarrow \langle Cost_{new}, T_{new} \rangle$
\State $P[i+m, j+n, u+k] \leftarrow $$(i, j, u, m, n, k)$
\EndIf
\EndIf
\EndFor
\EndFor
\Return Backtrack $P$ from $\arg\min_{u} DP[M, N, u].c$
    \end{algorithmic}  
\end{algorithm}

\subsection{Overall Performance}
\subsubsection{Throughput Evaluation}As shown in Fig. \ref{fig:exp2}, Diffusers and Dynamic Batching saturate prematurely due to poor hardware computation utilization and synchronization bottlenecks. While InstGenIE adopts continuous batching, it hits a ceiling as cumulative latency spikes throttle execution at high loads. In contrast, SynerDiff exhibits superior scalability, with a peak throughput of 3.4 tasks/s under burst traffic and outperforms InstGenIE by 1.6$\times$ at $\lambda=3.0$. This gain stems from our VAE Chunking and Skip-CFG strategies and feedback controller, which neutralize concurrent interference under dynamic loads. By achieving latency-lossless concurrency, SynerDiff fully unlocks the throughput potential of continuous batching.

\subsubsection{E2E Latency Evaluation}
\begin{figure}[t]
   \centering
        \setlength{\abovecaptionskip}{-3pt}
   \begin{minipage}[t]{0.24\textwidth}
     \centering
     \includegraphics[width=\textwidth]{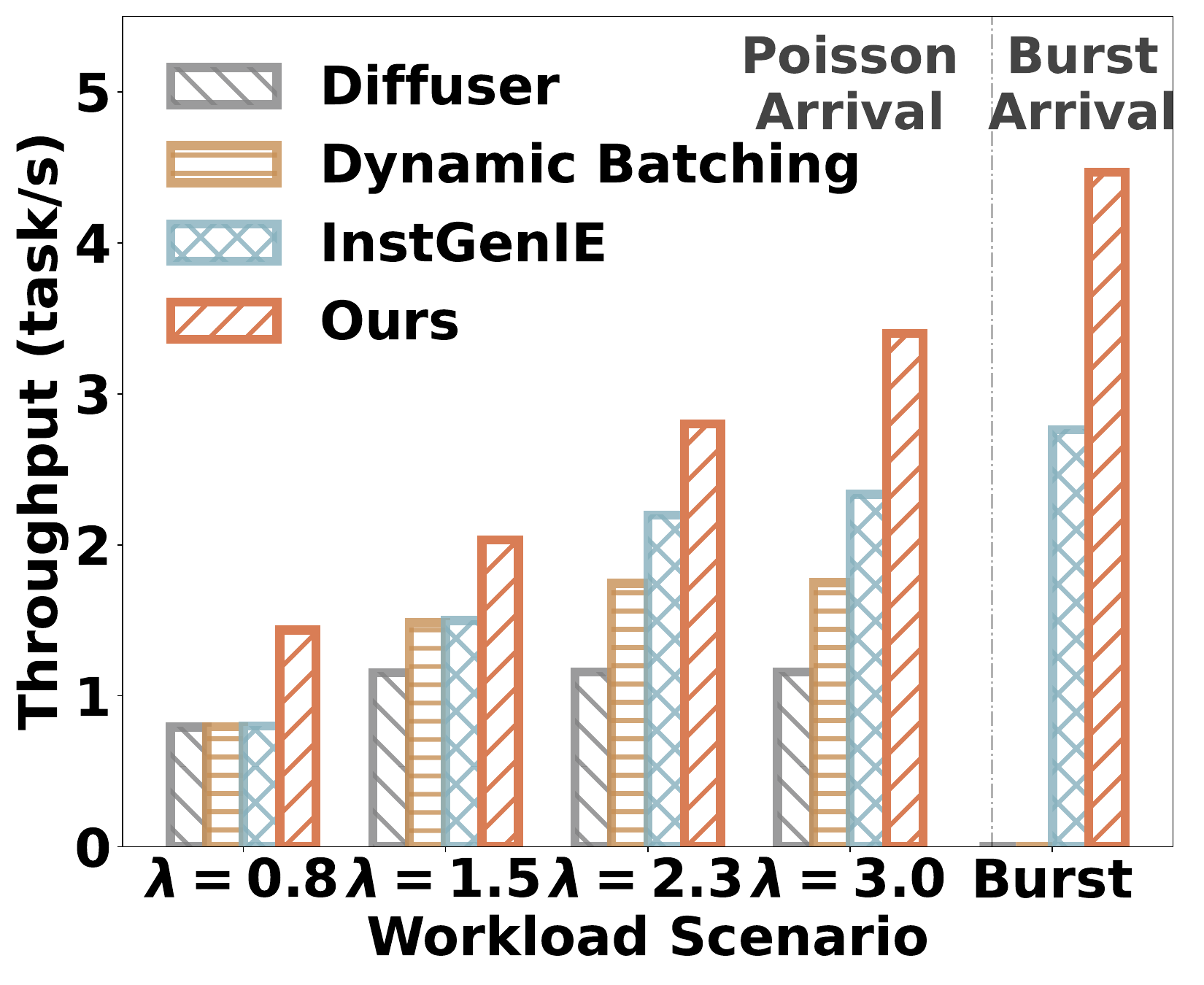}
     \caption{System throughput analysis.}
     \label{fig:exp2}
   \end{minipage}
   \hfill
   \begin{minipage}[t]{0.24\textwidth}
     \centering
     \includegraphics[width=\textwidth]{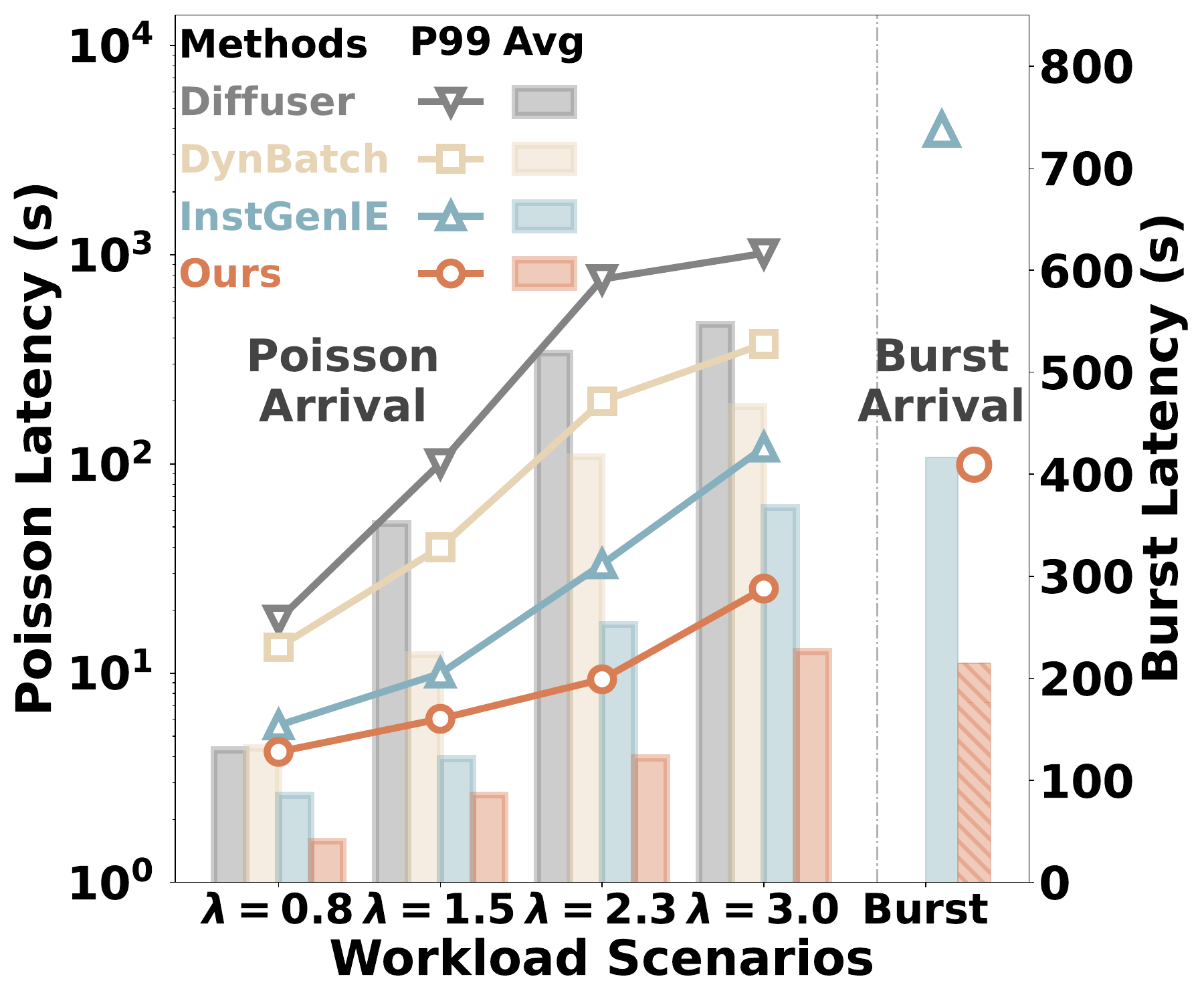}
     \caption{Average latency analysis.}
     \label{fig:exp3}
   \end{minipage}
     	\vspace{-0.3cm}
 \end{figure}

Fig. \ref{fig:exp3} illustrates E2E latency trends across loads. At $\lambda=2.3$, baselines suffer exponential spikes in average and P99 latencies as queue delays accumulate due to throughput saturation. In contrast, SynerDiff remains stable latency under extreme workloads, reducing latency by 78.7\% compared to InstGenIE at $\lambda=3.0$. This resilience stems from the synergy scheduling: the global scheduler plans concurrent sequences and configures VAE Chunking and Skip-CFG decisions to minimize average E2E latency, and the feedback controller adaptively elevates the throughput based on queue loads to mitigate tail latency, collectively ensuring the lowest average and P99 latencies under heavy workloads.


\subsubsection{Component Real-time Performance Evaluation}To uncover the mechanisms driving system gains, Fig. \ref{fig:exp4} analyzes UNet and VAE intra-concurrency performance. While InstGenIE suffers a 45\% throughput drop due to severe latency spikes, SynerDiff maintains stable UNet throughput—even outperforming non-concurrent steps during bursts. Furthermore, SynerDiff maintains stable VAE decoding latency, whereas InstGenIE’s surges under load. This efficiency stems from our intra-inter synergy: VAE Chunking and Skip-CFG neutralize interference to boost UNet throughput and the global scheduler plans task sequences to minimize average VAE latency, achieving high throughput and low task delay simultaneously under pressure.


\subsubsection{Generation Quality Assessment}
\begin{figure}[t]
        \setlength{\abovecaptionskip}{-3pt}
	\centering
	\includegraphics[width=1.0\linewidth]{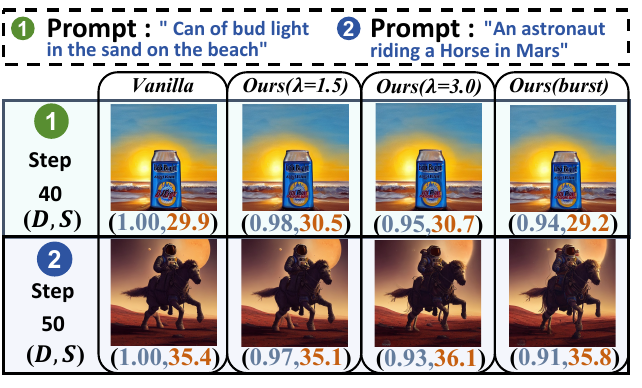}
	\vspace{-0.2in}
	\caption{Visualization of images generated by SynerDiff, where (D,S) denotes the DINO Score and CLIP Score, respectively.}
	\label{fig:exp1}
   	\vspace{-0.3cm}
\end{figure}

Table \ref{tab:2} evaluates SynerDiff’s image fidelity and text alignment across varying arrival rates. The DINO score remains robust, decreasing only 13\% even under burst arrival as the pre-defined Skip-CFG floor prevents the global scheduler from skipping sensitive early denoising stages, averting structural distortions. Visualizations in Fig. \ref{fig:exp1} further confirm minimal perceptual shifts and no structural discrepancies compared to vanilla. CLIP score of ours also remains comparable to ground truth. Overall, SynerDiff maintains high-fidelity, text-consistent generation while significantly boosting throughput and reducing latency.

 \begin{table}[t]
  \renewcommand{\arraystretch}{1.2}
 \small
    \centering
    \caption{Average quality of generated images.} \label{tab:2}
    \vspace{-0.2cm}
\resizebox{0.5\textwidth}{!}{
\centering
\begin{tabular}{c | c c c c c }
       \toprule
    \textbf{Model}&\textbf{Vanilla}&\textbf{Ours ($\lambda$=1.5)}&\textbf{Ours ($\lambda$=2.3)}&\textbf{Ours ($\lambda$=3.0)}&\textbf{Ours (burst)}   \\\cline{1-6}
        DINO Score&1&0.953&0.926&0.91&0.872 \\ \cline{1-6}
        CLIP Score&32.68&32.52&32.41&32.41&32.39 \\ 
       \bottomrule
    \end{tabular} }
\vspace{-0.3cm}
\end{table}

\subsection{Ablation study}

\begin{figure}[t]
   \centering
        \setlength{\abovecaptionskip}{-3pt}
   \begin{minipage}[t]{0.24\textwidth}
     \centering
     \includegraphics[width=\textwidth]{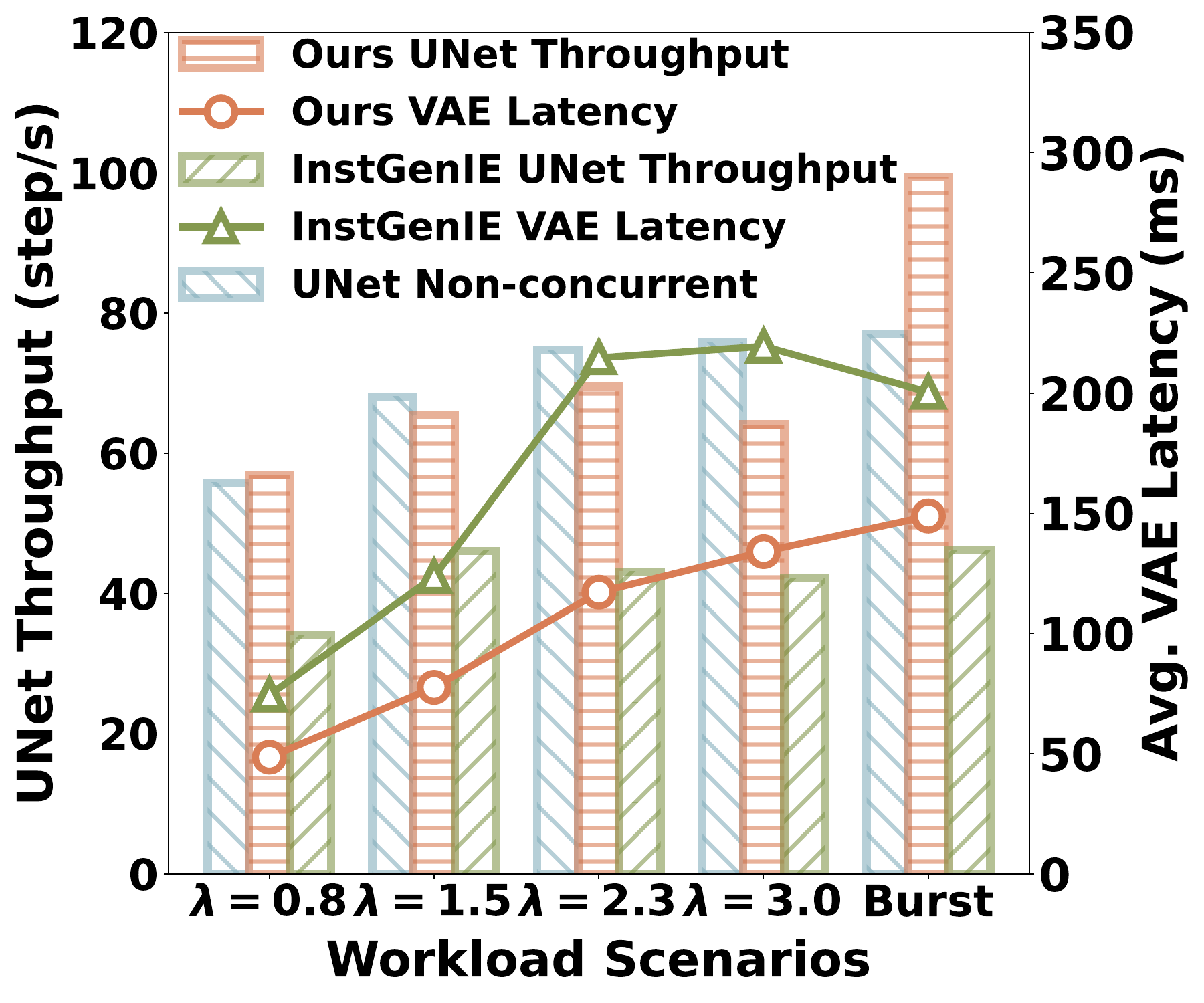}
     \caption{Intra-concurrency level analysis of UNet and VAE.}
     \label{fig:exp4}
   \end{minipage}
   \hfill
   \begin{minipage}[t]{0.24\textwidth}
     \centering
     \includegraphics[width=\textwidth]{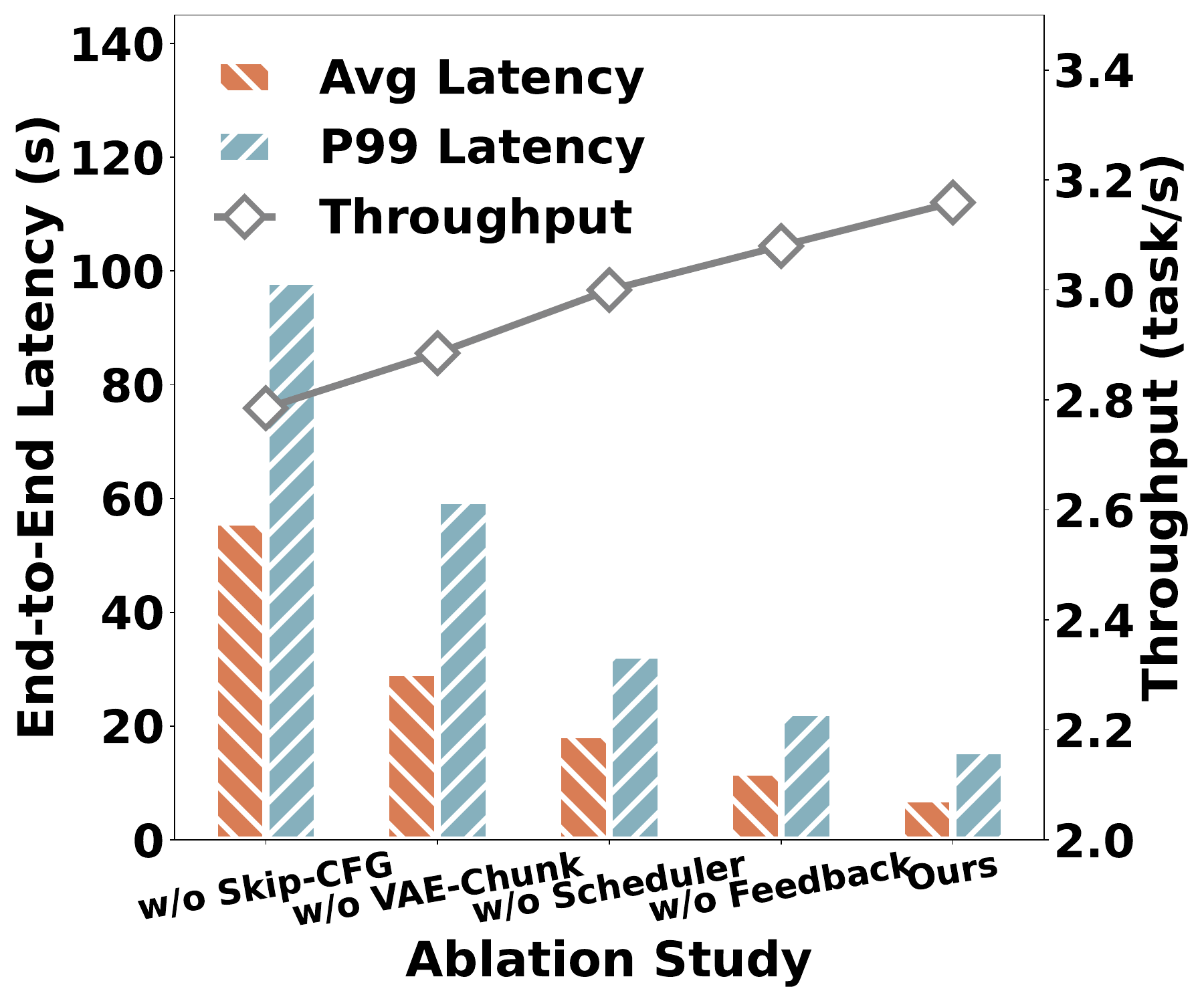}
     \caption{Ablation study of SynerDiff components on system performance.}
     \label{fig:exp5}
   \end{minipage}
     	\vspace{-0.3cm}
 \end{figure}

Ablation results in Fig. \ref{fig:exp5} demonstrate that Skip-CFG and VAE Chunking are fundamental to performance, as their absence reduces throughput by 20\% and inflates average task latency by 3.07$\times$. Furthermore, the global scheduler and feedback controller contribute an additional 60\% reduction in E2E latency and a 10\% throughput boost. This highlights a clear synergy among these components: while intra-concurrency strategies mitigate resource contention for near-lossless concurrency, the inter-concurrency scheduling achieves globally optimal, load-adaptive task scheduling. Ultimately, these synergies achieve a cumulative 87\% latency reduction, establishing SynerDiff as a highly efficient and scalable paradigm for DM serving.


\section{Conclusion}

In this paper, we present SynerDiff, a continuous batching system optimizing concurrent UNet and VAE execution to achieve high throughput and low E2E latency for DM serving. At the intra-concurrency level, SynerDiff eliminates resource-induced latency spikes by integrating VAE Chunking and Skip-CFG. At the inter-concurrency level, SynerDiff employs a threshold-aware scheduler to plan multi-task execution sequences and tune intra-concurrency configurations, minimizing task latency while sustaining high-level throughput under varying workloads. Experimental results validate that SynerDiff achieves a 1.6$\times$ throughput boost and reduces average E2E latency by up to 78.7\%, while preserving image fidelity. In future work, we will explore the parallel scalability of SynerDiff across multi-GPU environments and investigate its integration with single-task optimization techniques to further elevate the service efficiency of diffusion model batching system.



\bibliographystyle{IEEEtran}
\bibliography{ref}

\vspace{12pt}

\end{document}